\documentclass[10pt,twocolumn,letterpaper]{article}

\usepackage{wacv}
\usepackage{times}
\usepackage{epsfig}
\usepackage{graphicx}
\usepackage{amsmath}
\usepackage{amssymb}
\usepackage{multirow}
\usepackage{microtype}
\usepackage{booktabs}
\usepackage{makecell}
\usepackage{caption} 
\def\wacvPaperID{28} 

\wacvfinalcopy 
\ifwacvfinal
\usepackage[breaklinks=true,bookmarks=false]{hyperref}
\else
\usepackage[pagebackref=true,breaklinks=true,colorlinks,bookmarks=false]{hyperref}
\fi

\begin{document}

\title{Improving Person Re-Identification with Temporal Constraints}

\author{Julia Dietlmeier*, Feiyan Hu*, Frances Ryan, Noel E. O'Connor, Kevin McGuinness\\
Insight SFI Research Centre for Data Analytics\\
Dublin City University, Glasnevin 9, Dublin, Ireland\\
{\tt\small \{julia.dietlmeier,feiyan.hu,frances.ryan,noel.oconnor,kevin.mcguinness\}@insight-centre.org}\\
* \small{All these authors contributed equally}

}

\maketitle

\begin{abstract}
   In this paper we introduce an image-based person re-identification dataset collected across five non-overlapping camera views in the large and busy airport in Dublin, Ireland. Unlike all publicly available image-based datasets, our dataset contains timestamp information in addition to frame number, and camera and person IDs. Also our dataset has been fully anonymized to comply with modern data privacy regulations. We apply state-of-the-art person re-identification models to our dataset and show that by leveraging the available timestamp information we are able to achieve a significant gain of {\bfseries37.43}\% in mAP and a gain of {\bfseries30.22}\% in Rank1 accuracy. We also propose a Bayesian temporal re-ranking post-processing step, which further adds a {\bfseries10.03}\% gain in mAP and {\bfseries9.95}\% gain in Rank1 accuracy metrics. This work on combining visual and temporal information is not possible on other image-based person re-identification datasets. We believe that the proposed new dataset will enable further development of person re-identification research for challenging real-world applications. 
\end{abstract}

\section{Introduction}
\label{sec:intro}
Person re-identification (re-ID) aims at matching people across multiple non-overlapping camera views. Specifically, given a \textit{query} image the task is to find a match in a large \textit{gallery} of images. This task generally falls into the category of image retrieval problems. The environment in which the real-world person re-ID systems are deployed ranges from university campuses and streets to airports and train stations~\cite{Dietlmeier_ICPR2020}. Currently, deep learning-based person \text{re-ID} has reached performance saturation on several public datasets. An example of this is the Market1501 dataset where some methods achieve higher Rank1 accuracy than humans \cite{Ye_SurveyTPAMI2021}. On the other hand, the performance on the challenging VIPeR dataset still remains comparatively low. Furthermore, person re-ID faces several key challenges such as various camera angles, varying lighting conditions and occlusions \cite{Meng_CVPR2019_WeaklyReID}. The so-called ``appearance ambiguity'' \cite{Wang_AAAI2019} problem, where different individuals may have a similar appearance, makes it difficult for most person re-ID models to improve performance further by focusing solely on visual features.

Modern person re-ID research concentrates on deep learning-based approaches and over time migrated from the handcrafted features and distance metric learning. A timely and comprehensive review of the re-ID methods is provided in \cite{Ye_SurveyTPAMI2021}. The size of publicly available academic person re-ID datasets has increased over time and this has facilitated the training of large deep learning-based models. Most present person re-ID methods, however, still do not scale to large datasets and ignore spatial-temporal constraints that could potentially alleviate the appearance ambiguity problem. The core of these algorithms is to learn  pedestrian features and similarity functions that are view invariant and robust to camera change \cite{Lv_CVPR2018}. These methods are still far from applicable to large-scale real-world scenarios containing a large amount of candidate images (which, following the literature, we refer to as gallery images) that need to be searched to re-identify a person \cite{Wang_AAAI2019,Cho_CVIU2019}. An airport is a good example of this scenario, where there are potentially thousands of cameras, each generating tens of thousands if not hundreds of thousands of frames per minute, all of which are potential candidates for re-identification should spatial-temporal priors not be taken into account. The re-ID research on fusing visual and spatial-temporal information is, however, being hindered by the availability of suitable datasets. In this work, we make three contributions:
\begin{table*}
\centering
  \begin{tabular}{lllllll}
    \toprule
    Name & Cameras & Query & Gallery  & Frame \# & Time & Anonymized \\
    \midrule
    Airport \cite{Karanam_IEEETPAMI2018,Camps_IEEECSVT2017} & 6 & 1,003 & 2,420 & yes & no & no \\
    Market1501 \cite{Zheng_ICCV2015_Market1501} & 6 & 3,368 & 19,732 & yes & no & no \\
    DukeMTMC-reID \cite{Gou_CVPR2017_DukeMTMC} & 8  & 2,228 & 17,661 & yes & no & no \\
    CUHK03 \cite{Li_CVPR2014_CUHK03}           & 6  & 5,332 & 1,400  & no  & no & no \\
    VIPeR \cite{Gray_ECCV2008_ViPER}           & 2  & 316   & 316    & no  & no & no \\
    DAA (Ours) & 5 & 71 & 1,029 & yes & yes & yes \\
    \bottomrule
  \end{tabular}
  \caption{Comparison of commonly used image-based person re-ID datasets.}
  \label{tab:datasets}
\end{table*}
\begin{itemize}
  \item First, we introduce a new dataset, called DAA\footnote{The name DAA stands for Dublin Airport Authority}, collected in the large and busy Dublin Airport in Ireland. The main feature of the DAA dataset is that all images contain time information (timestamps) in addition to the frame number, person, and camera IDs. Also, our dataset has been fully anonymized to comply with the strict data privacy regulations. 
  
  \item Second, we show that by leveraging the available temporal information we are able to significantly increase the performance of the state-of-the-art person re-ID methods selected. Specifically, we achieve a substantial gain in mAP of {\bfseries37.43}\% and a Rank1 gain of {\bfseries30.22}\%.
  
  \item Third, we develop an efficient temporal re-ranking algorithm that further adds {\bfseries10.03}\% gain in mAP and {\bfseries9.95}\% gain in Rank1 accuracy metrics.
\end{itemize}

The remainder of the paper is organized as follows:  Section~\ref{related_work} reviews the commonly used image-based person re-ID datasets. Section~\ref{sec:new_dataset} describes our new dataset and its key features. Section~\ref{sec:methodology} outlines the methodology and experiments. Section~\ref{sec:re-ranking} introduces a new temporal re-ranking step and  Section~\ref{sec:discussion} discusses our findings.

\section{Related Work}
\label{related_work}

Person re-ID datasets can be mainly categorized into image- and video-based \cite{Ye_SurveyTPAMI2021,Zheng_PastPresentFuture_arXiv2016}. In this work we focus on the first category where the query person is represented by an image (extracted from a CCTV video). Below, we review the image-based datasets commonly used in the person re-ID community in the chronological order in which they were released. Table~\ref{tab:datasets} details the image-based datasets reviewed in this section.

\begin{table*}
\centering
  
\begin{tabular}{llccc|ccc}
\toprule
&  & \multicolumn{3}{c}{Market1501} & \multicolumn{3}{c}{Proposed DAA dataset} \\
&  & mAP  & Rank1  & Rank5  & mAP   & Rank1  & Rank5   \\
\midrule
\multirow{4}{*}{MLFN} & Original &  74.3 & 90.1 & 95.9 & 35.9 &66.2 & 88.7\\
& +TR    & $+2.2$ & $+1.2$ & $+1.0$ & $+6.8$ & 0      & $-5.6$ \\
& +TLift & $+1.2$ & $+2.0$ & $+1.0$ & $+4.9$ & $+1.4$ & $-4.2$ \\
& +TLift*& $+1.0$ & $+1.5$ & $+0.8$ & $+7.1$ & $-2.8$ & $-4.2$ \\
\midrule
\multirow{4}{*}{HACNN} & Original &  75.3 & 90.6 & 96.2 & 31.8 &62.0 & 83.1\\
& +TR    & $+2.2$ & $+1.6$ & $+0.5$ & $+4.6$ & 0      & $-5.6$ \\
& +TLift & $+1.3$ & $+1.7$ & $+0.9$ & $+3.9$ & $-1.4$ & $-1.4$ \\
& +TLift*& $+1.2$ & $+1.6$ & $+0.7$ & $+7.1$ & $-1.4$ & $-4.2$ \\
\bottomrule
\end{tabular}
    \caption{Performance increase with the introduction of frame numbers on Market1501 and our proposed datasets. TR (Temporal Re-ranking) is our proposed re-ranking approach using temporal prior in Section~\ref{sec:re-ranking}. TLift, proposed in~\cite{liao2020interpretable},  models temporal probability with a Gaussian Kernel. TLift* is the same method as TLift, but uses the Gamma distribution instead.}
    \label{tab:compare_market}
\end{table*}

The Viewpoint Invariant Pedestrian Recognition (\textbf{VIPeR}) \cite{Gray_ECCV2008_ViPER} dataset was released in 2008 and contains 632 person images from two disjoint outdoor camera views. The dataset was collected under different viewpoints and illumination conditions and therefore is considered as one of the most challenging.
Each person has one image per camera. Each image is resized to be 48 $\times$ 128 pixels. In addition, each image has a pedestrian viewpoint information tag of 0° (front), 45°, 90° (right), 135°, and 180° (back) degrees. 
The \textbf{CUHK03} \cite{Li_CVPR2014_CUHK03} dataset was collected from six surveillance cameras in the University campus environment and released in 2014. 
It was the first large-scale dataset that was large enough for training deep learning-based re-ID models. Each person identity is captured by two disjoint camera views and has an average of 4.8 images in each view. 
Because it was collected in a more realistic setting, this dataset features such problems as misalignment, occlusions, and missing body parts.
The bounding boxes were extracted using manual annotations and the Deformable Part Models (DPM) object detector \cite{Felzenszwalb_TPAMI2010_DPM}. 
The next very popular large-scale dataset, \textbf{Market1501} \cite{Zheng_ICCV2015_Market1501}, was released in 2015. 
It also features images extracted using the DPM detector and thus inherits problems of background clutter and misalignment. 
Other problems include viewpoint variations, detection errors, and low resolution \cite{Karanam_IEEETPAMI2018}. The dataset was collected from a total of six cameras (overlapping exists) placed in front of a campus supermarket in Tsinghua University. 
The authors provide false positive bounding boxes and also 2,793 false alarms and 500,000 distractors. There are 3.6 images on average for each of 1,501 identities and the total number of images amounts to 32,668. 
\textbf{DukeMTMC-reID} \cite{Gou_CVPR2017_DukeMTMC} is another large-scale dataset introduced in 2017 that features images collected outdoors on the Duke University campus.
The dataset contains manually labeled bounding boxes for a total of 1,812 identities from eight cameras. There are 1,404 identities appearing in more than two cameras and 408 identities (distractors) appearing in only one camera \cite{Zheng_DukeMTMC_arXiv2017}. Further, the dataset contains 2,228 query images, 17,661 gallery images, and 16,522 training images.
\textbf{Airport} \cite{Camps_IEEECSVT2017,Karanam_IEEETPAMI2018} is a relatively recent large-scale image-based person re-ID dataset released in 2018 that is close to our work in both aspects: a real-world airport environment and associated image type and quality. The data were collected from six non-overlapping indoor surveillance cameras in a mid-sized airport in Cleveland, USA. Each camera captured 12-hour long videos at 30 frames per second from 8AM to 8PM. In total, there are 39,902 bounding boxes and 9,651 unique identities. The authors employed the ACF \cite{Dollar_TPAMI2014_ACF} framework to extract bounding boxes. Although not elaborated in \cite{Camps_IEEECSVT2017} and \cite{Karanam_IEEETPAMI2018}, the images in the Airport dataset carry some relative time information. An example of an image from the query folder is \texttt{00011004\_c37\_t01\_0002.jpg}. From here we infer that the time is not given in seconds as $t$ ranges from 1 to 40 in the query and gallery folders. We presume that this is the video clip number as each video was split in 40 clips (each 5 minutes long)~\cite{Karanam_IEEETPAMI2018}. We could not find any publication exploiting this temporal characteristic of the Airport dataset. In contrast to the previous datasets that capture data in public settings, the video data in the Airport dataset is streamed from the inside of the secure area, beyond the security checkpoint, of an airport \cite{Karanam_IEEETPAMI2018}. As highlighted by the authors, it is generally very difficult to obtain data from such a camera network, which is similar to ours.

A few papers on spatial-temporal person re-ID \cite{Li_2020_STreranking,Lv_CVPR2018,Wang_AAAI2019, liao2020interpretable} work with DukeMTMC-reID, Market1501, and sometimes the small-scale GRID dataset \cite{GRID_CVPR2009}. In these works a common approach is to treat the frame number as timestamp. The evaluation of the temporal performance of re-ID models is still considered to be under explored \cite{Karanam_IEEETPAMI2018,Wang_AAAI2019}.

In the aforementioned datasets, which contain incomplete temporal information such as frame numbers, relative temporal difference within a camera could be inferred. Temporal differences across cameras, however, are impossible to infer. To demonstrate the extra performance that knowledge of frame numbers could impart, Table~\ref{tab:compare_market} shows a performance increase with the introduction of frame numbers on Market1501 and the proposed DAA dataset. We benchmark two state-of-the-art re-ID models such as MLFN~\cite{chang2018multi} and HACNN~\cite{li2018harmonious} and observe that using the provided frame numbers can boost performance, especially mAP. Even by just using time prior (Section~\ref{sec:re-ranking}) within each camera, performance is better than that achieved with more complicated ranking methods like TLift~\cite{liao2020interpretable}. The frame number can only be used to compute the within-camera time difference; however, most applications would require cross-camera search. To compute cross-camera time differences, timestamp information is needed instead of frame numbers. In Section~\ref{sec:re-ranking}, we show that significant improvements can be made just by using the timestamp information included in the proposed DAA dataset to re-rank results. This is shown in Table~\ref{tab:second_exp} and Table~\ref{tab:reranking_results}.

We conclude that none of the datasets reviewed above explicitly provide timestamp information and none have been anonymized (see Table~\ref{tab:datasets}). It is now accepted that there are significant privacy concerns when releasing new datasets across research communities \cite{malin2011never}. At the time of writing, some of the image-based person re-ID datasets listed in Table~\ref{tab:datasets} have been taken offline due to the privacy concerns \cite{Murgia_FT2019} and/or facing potential legal challenges. It is reasonable to envisage that anonymization of faces could lead to performance decaying. However, recent work \cite{Dietlmeier_ICPR2020} shows that face anonymization of person re-ID datasets only marginally affects the performance of state-of-the-art person re-ID methods. Based on these findings we release an anonymized version of this new person re-ID dataset.

\section{New Dataset}
\label{sec:new_dataset}

\begin{figure*}
  \centering
  \includegraphics[width=\linewidth]{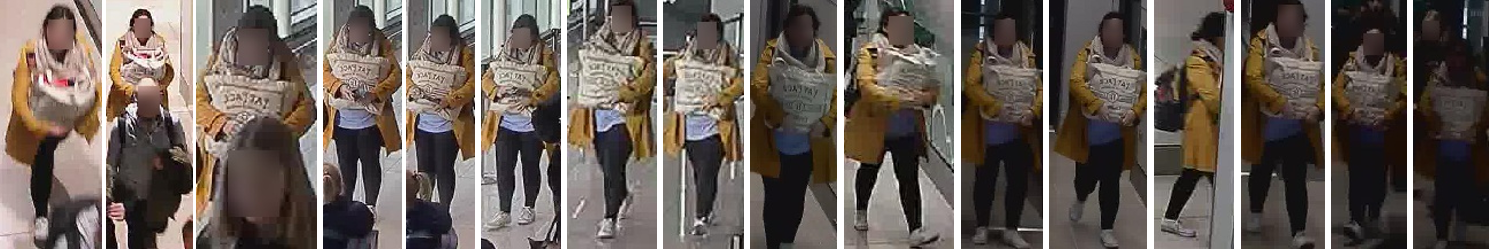}
  \caption{Examples of anonymized gallery images for the person identity ID=2708. It can be seen that our dataset has such challenging attributes as occlusions, viewpoint variations, pose and illumination variations, missing body parts and background clutter.}
  \label{fig:gallery_sample}
\end{figure*}
\begin{figure}[ht]
{\includegraphics[width=1\linewidth]{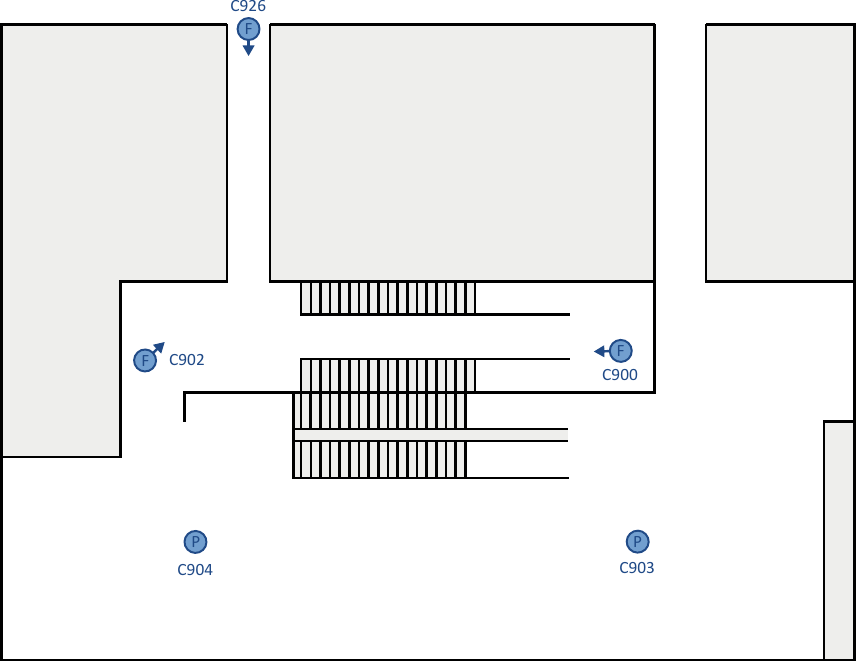}}
\caption{Camera topology of our DAA dataset}
\label{fig:topo_and_map}
\end{figure}

The proposed DAA dataset was collected in one day from 14-hours long CCTV surveillance videos starting from 3AM to 5PM in one transfer level of the busy airport in Dublin, Ireland. The camera network in this transfer level consists of five digital cameras covering non-overlapping fields of view and streaming videos at variable frame rates. Varying illumination conditions, reflective floor, and reflections in windows and on the walls make up some of the challenges for this work. Figure~\ref{fig:topo_and_map} shows the camera topology of the dataset;  Table~\ref{tab:walkingdistances} provides the relative walking distances between cameras.

We extract and tag the timestamp information (in seconds) when processing videos and extracting keyframes with \textit{ffmpeg}. The inclusion of time information is the most prominent feature of our dataset. We encode metadata about the images in the filenames; an example can be given by \texttt{21\_c0900\_t36000\_frame0002300\_2.jpg}. Here the person ID is 21, and the camera ID is \texttt{c0900}. The \texttt{t36000} code means this image was sampled at 10AM so that $t = 10 \times 60 \times 60 = 36,000$ seconds. The frame number is 2300 and this is the second bounding box extracted from that frame. 

We perform annotation at two levels. In the first-level annotation we extract bounding boxes containing people. The resolution of each camera is $1920 \times 1080$ pixels. For the query camera C900 we manually annotate the bounding boxes using the open source \textit{DarkLabel} tool. For all gallery cameras we use generic object detector SSD \cite{Liu_ECCV2016_SSD} (SSD512 model\footnote{https://github.com/pierluigiferrari/ssd\_keras}) with the confidence threshold set to 0.25 to automatically generate candidate bounding boxes from the keyframes extracted. Therefore, the sizes of cropped bounding boxes vary in our dataset. In the second-level annotation procedure we manually match people across the four gallery cameras C902, C903, C904, and C926. We had to manually process 195,978 bounding boxes in total. This was the most time-consuming process. The resulting query folder of our dataset consists of 71 unique identities from camera C900 and the resulting gallery folder contains 1,029 images from cameras C902, C903, C904, and C926 with one or more images per identity. Figure~\ref{fig:gallery_sample} illustrates the challenging attributes of our dataset (such as occlusions, viewpoint variations, pose and illumination variations, background clutter, and missing body parts) on a sample ID from the gallery folder. 

For the anonymization we adopt the procedure recommended in \cite{Dietlmeier_ICPR2020} using the pre-trained TinyFaces face detector \cite{Hu_CVPR2017_TinyFaces} to detect the faces first and then blur the regions detected with the large-kernel Gaussian to remove all privacy-sensitive information. We manually verify the quality of anonymization. 
Our DAA dataset is available under this link: \url{https://bit.ly/3AtXTd6}.


\section{Methodology}
\label{sec:methodology}
In this section we benchmark some state-of-the-art person re-ID models on our new dataset 
and show that by using the available temporal constraint we are able significantly to boost  performance.

\subsection{Performance metrics}

We evaluate the performance of re-ID models using the mean average precision (mAP) and Rank1 to Rank20 measures. These metrics rely on ranking images in the gallery dataset according to similarity with the query images. A similarity score is generated for each possible query/gallery image combination. For each query image, this similarity score is then used to rank all of the gallery images according to the estimated likelihood of the gallery image containing the individual in the query image. The top ranked image is the one that the system estimates as having the highest probability of containing the person in the query image, the second ranked image is the next most likely, and so on.
\begin{table*}
\centering
  
  \begin{tabular}{llccccc}
    \toprule
    & Model & mAP & Rank1  & Rank5 & Rank10 & Rank20\\
    \midrule
    1 & PCB~\cite{Sun_ECCV2018_PCB}    & 33.3$\pm$1.6 & 63.8$\pm$4.5 & 83.5$\pm$3.3 & 88.8$\pm$2.2 & 93.7$\pm$1.8\\
    2 & MLFN~\cite{chang2018multi}  & \textbf{38.0$\pm$1.6} & 66.7$\pm$3.3 & 86.9$\pm$2.8 & 91.7$\pm$1.7 & 94.5$\pm$0.9\\
    3 & HACNN~\cite{li2018harmonious}  & 31.6$\pm$0.9 & 65.4$\pm$2.9 & 82.7$\pm$2.1 & 88.7$\pm$2.1 & 93.9$\pm$1.5\\
    4 & Resnet50Mid~\cite{yu2017devil} & 32.9$\pm$2.0 & 65.7$\pm$4.7 & 83.5$\pm$2.9 & 89.2$\pm$1.6 & 92.7$\pm$2.3\\
    5 & MuDeep~\cite{qian2017multi} & 18.8$\pm$0.4 & 52.3$\pm$2.5 & 74.2$\pm$2.3 & 81.8$\pm$2.2 & 89.2$\pm$1.3 \\
    6 & BoT~\cite{Luo2019BoT} & 37.1$\pm$1.8 & \textbf{72.3$\pm$4.4} & \textbf{89.5$\pm$2.6} & \textbf{93.8$\pm$2.3} & \textbf{96.4$\pm$1.5} \\
  \bottomrule
\end{tabular}
  \caption{Benchmarks on our DAA re-ID dataset. Training for each model is repeated 20 times, and mean performance and standard deviation are reported.}
  \label{tab:benchmark}
\end{table*}
Comparing the ranked results with the ground truth (images known to contain the same person as the query image) allows the computation of several metrics. The Rank1 metric evaluates the proportion of queries where the top-ranked gallery image contains the correct person (i.e.~the same person appeared in both the query image and the top ranked gallery image); the Rank5 metric evaluates the proportion of queries where at least one of the top-5 ranked gallery images contains the correct person; and Rank10 and Rank20 are the proportion of queries identifying the correct person in at least one of the top 10 or 20 ranked gallery images, respectfully. Mean average precision (mAP) is calculated by taking the mean of average precision (AP) scores across all queries.
It gives an estimate of how many images would contain the correct person in the top-$n$ images for all $n$. In the evaluation of experiments we use a similar approach as in Market1501: gallery images, that share their person ID and camera ID with a query image, are excluded from evaluation.

%
\subsection{Experiments}

We conduct benchmark testing on the proposed DAA dataset. Table~\ref{tab:benchmark} details the results of this experiment. We use the Torchreid library by Kaiyang Zhou and Tao Xiang \cite{KaiyangZhou_TorchReID_arXiv2019} to implement the state-of-the-art models 1 to 5. We adopt a cross-dataset methodology and train each model on the Airport \cite{Camps_IEEECSVT2017,Karanam_IEEETPAMI2018} dataset (3,493 images) for 60 epochs and evaluate on our DAA dataset. We train the BoT model\footnote{https://github.com/michuanhaohao/reid-strong-baseline} for 120 epochs on the Airport dataset. Experiments for each model are repeated 20 times. We report the mean and standard deviation of each metric in these repeated experiments. Table~\ref{tab:benchmark} shows that MLFN and BoT are the two best models.

One thing to note when considering the results is that the \text{re-ID} work is skewed towards individuals wearing distinctive, often brightly coloured, clothing. Annotation, even manually, of individuals wearing dark clothing can be very challenging. Although ideally we would like to have completed our performance evaluation on a more balanced dataset of individuals, in the context of Dublin Airport, being able to correctly re-identify a few individuals within a crowd and combine this with crowd density mapping has the potential to provide information about general passenger flow through a space, as timestamps at different cameras 
can be used to pinpoint their appearance in a certain area at a given time.

Next, we proceed eliminating irrelevant gallery images using time information. We achieve the first improvement in our re-ID performance by reducing the number of images that are considered when searching the dataset to re-identify a particular person. Our full annotated dataset contains 1,029 candidate images in total for 71 identities. Considering all of these images for every identity being searched can result in an increased potential for false positives, which can arise, for example, by two individuals appearing very similar by coincidence. By imposing time constraints on gallery images we also reduce the total number of images to be processed and reduce theoretical number of multiplications needed when computing distances between query-gallery pairs. 


Clearly, there is a strong prior over the candidate images in the gallery that arises from the fact that temporally close images are more likely to contain the same people. Specifically, people entering the airport for the first time only re-appear in future frames, and after leaving one camera are more likely to appear in another camera within a short period of time. We propose to constrain the search based on the available timestamp information~$t$. In this experiment on our DAA dataset we construct reduced galleries for each query based on $t_\text{query} + \Delta t$ where $\Delta t = 30$ minutes, for example. This step is based on the observation that a particular person will usually pass from the query camera C0900 through the relevant gallery cameras C0902, C0903, C0904, and C0926 in less than 30 minutes. This approach follows the procedure mentioned in \cite{Wang_AAAI2019}, where authors use a spatial-temporal constraint to eliminate lots of irrelevant images in the gallery.

After constructing reduced galleries for each of the 71 query images, we run each re-ID model 71 times and compute the aggregate statistics afterwards. Table~\ref{tab:second_exp} shows the results for $T_{min} \leq \Delta t \leq T_{max}$ minutes, clearly demonstrating the enormous improvement in mAP and Ranks 1 to 10. Specifically, we report the average performance gain in mAP across all models for the time interval $[0,30)$ as \textbf{37.43}$\%$ and \textbf{30.22}$\%$ gain in Rank1 accuracy. 

\newcommand{\bla}{blah blah blah blah blah blah blah blah}

\newcommand{\pcbAA}{47.2$\pm$0.5 \textbf{91.6$\pm$1.6} 96.4$\pm$0.7 97.1$\pm$0.3}
\newcommand{\mlfnAA}{48.3$\pm$0.5 89.2$\pm$2.6 96.3$\pm$0.9 97.2$\pm$0.0}
\newcommand{\haAA}{46.5$\pm$0.4 92.2$\pm$1.8 96.2$\pm$0.8 97.1$\pm$0.3}
\newcommand{\resAA}{46.0$\pm$0.7 90.6$\pm$2.5 96.3$\pm$0.8 97.1$\pm$0.3}
\newcommand{\mdAA}{40.5$\pm$0.5 80.4$\pm$2.5 \textbf{93.7$\pm$1.3} \textbf{96.3$\pm$0.9}}
\newcommand{\botAA}{47.5$\pm$0.9 93.3$\pm$1.8 96.6$\pm$0.8 97.2$\pm$0.0}

\newcommand{\pcbA}{71.7$\pm$0.8 91.1$\pm$1.6 \textbf{97.4$\pm$1.0} \textbf{98.5$\pm$0.4}}
\newcommand{\mlfnA}{74.3$\pm$0.8 \textbf{89.7$\pm$2.3} \textbf{97.0$\pm$1.1} \textbf{98.2$\pm$0.6}}
\newcommand{\haA}{70.9$\pm$0.6 \textbf{93.2$\pm$1.6} \textbf{97.3$\pm$0.8} \textbf{98.5$\pm$0.3}}
\newcommand{\resA}{\textbf{70.0$\pm$1.4} \textbf{90.7$\pm$2.7} \textbf{96.6$\pm$1.2} \textbf{98.0$\pm$0.7}}
\newcommand{\mdA}{\textbf{57.5$\pm$0.7} \textbf{80.9$\pm$3.0} 92.0$\pm$1.7 95.5$\pm$1.3}
\newcommand{\botA}{73.2$\pm$1.3 \textbf{94.7$\pm$2.3} \textbf{97.8$\pm$1.1} \textbf{98.4$\pm$0.5}}

\newcommand{\pcbB}{\textbf{71.8$\pm$0.9} 90.8$\pm$1.8 {96.8$\pm$1.3} 98.2$\pm$0.6}
\newcommand{\mlfnB}{\textbf{74.5$\pm$0.8} 89.4$\pm$2.2 \textbf{97.0$\pm$1.1} \textbf{98.2$\pm$0.6}}
\newcommand{\haB}{\textbf{71.1$\pm$0.6} \textbf{93.2$\pm$1.6} \textbf{97.3$\pm$0.8} \textbf{98.5$\pm$0.3}}
\newcommand{\resB}{69.7$\pm$1.5 90.1$\pm$2.7 96.5$\pm$1.3 \textbf{98.0$\pm$0.7}}
\newcommand{\mdB}{56.8$\pm$0.7 80.7$\pm$3.2 91.5$\pm$1.8 95.4$\pm$1.5}
\newcommand{\botB}{\textbf{73.4$\pm$1.4} 94.5$\pm$2.2 \textbf{97.8$\pm$1.1} \textbf{98.4$\pm$0.5}}

\newcommand{\pcbC}{69.7$\pm$1.1 89.9$\pm$1.9 96.7$\pm$1.3 98.2$\pm$0.6}
\newcommand{\mlfnC}{73.1$\pm$0.9 88.3$\pm$2.2 95.9$\pm$0.9 97.9$\pm$0.8}
\newcommand{\haC}{68.8$\pm$0.6 92.3$\pm$1.8 96.3$\pm$0.9 \textbf{98.5$\pm$0.4}}
\newcommand{\resC}{67.8$\pm$1.6 89.0$\pm$2.8 96.3$\pm$1.1 97.6$\pm$0.6}
\newcommand{\mdC}{53.5$\pm$0.8 78.0$\pm$3.0 90.6$\pm$1.7 95.0$\pm$1.7}
\newcommand{\botC}{71.2$\pm$1.5 93.1$\pm$2.4 97.6$\pm$1.0 98.3$\pm$0.8}

\newcommand{\pcbD}{68.4$\pm$1.4 86.8$\pm$2.5 96.1$\pm$1.6 98.1$\pm$0.8}
\newcommand{\mlfnD}{72.4$\pm$0.9 86.1$\pm$2.4 95.1$\pm$0.8 97.7$\pm$0.8}
\newcommand{\haD}{67.6$\pm$0.6 89.2$\pm$2.4 96.0$\pm$1.0 97.6$\pm$0.6}
\newcommand{\resD}{67.1$\pm$1.7 86.7$\pm$3.1 95.4$\pm$2.0 97.4$\pm$0.8}
\newcommand{\mdD}{49.1$\pm$0.9 73.0$\pm$3.6 88.5$\pm$2.0 92.5$\pm$1.6}
\newcommand{\botD}{71.0$\pm$1.7 90.5$\pm$2.2 96.9$\pm$1.1 98.1$\pm$0.8}

\newcommand{\pcbE}{64.1$\pm$1.7 85.3$\pm$2.8 95.1$\pm$1.7 97.7$\pm$0.9}
\newcommand{\mlfnE}{68.2$\pm$1.1 85.4$\pm$2.7 94.4$\pm$1.3 97.5$\pm$1.0}
\newcommand{\haE}{62.8$\pm$0.8 85.9$\pm$2.8 95.8$\pm$1.3 97.6$\pm$0.6}
\newcommand{\resE}{62.7$\pm$1.6 84.6$\pm$2.9 95.2$\pm$2.1 97.2$\pm$0.9}
\newcommand{\mdE}{42.8$\pm$0.9 68.6$\pm$3.7 87.6$\pm$1.9 91.2$\pm$1.7}
\newcommand{\botE}{66.8$\pm$1.7 88.1$\pm$2.6 96.9$\pm$1.1 98.0$\pm$0.8}

\newcommand{\pcbEE}{46.5$\pm$1.6 76.4$\pm$3.8 90.9$\pm$1.9 94.4$\pm$1.8}
\newcommand{\mlfnEE}{50.5$\pm$1.4 77.1$\pm$3.1 91.0$\pm$1.7 93.7$\pm$1.5}
\newcommand{\haEE}{44.2$\pm$1.0 74.4$\pm$3.3 89.1$\pm$1.8 94.2$\pm$1.1}
\newcommand{\resEE}{44.7$\pm$2.0 74.2$\pm$5.3 89.6$\pm$2.2 93.4$\pm$2.1}
\newcommand{\mdEE}{25.0$\pm$0.9 51.0$\pm$4.0 72.3$\pm$2.9 80.6$\pm$2.5}
\newcommand{\botEE}{50.0$\pm$1.9 80.4$\pm$4.0 93.7$\pm$2.2 96.6$\pm$1.6}

\newcommand{\pcbF}{68.8$\pm$1.0 89.5$\pm$1.9 97.0$\pm$0.9 97.7$\pm$0.7}
\newcommand{\mlfnF}{71.1$\pm$0.8 88.7$\pm$2.8 96.3$\pm$0.8 97.0$\pm$0.7}
\newcommand{\haF}{67.4$\pm$0.5 91.9$\pm$1.6 96.6$\pm$1.2 97.7$\pm$0.9}
\newcommand{\resF}{66.1$\pm$1.4 89.1$\pm$2.3 95.6$\pm$1.7 97.3$\pm$1.2}
\newcommand{\mdF}{52.7$\pm$0.8 77.3$\pm$3.1 88.0$\pm$2.0 92.7$\pm$1.6}
\newcommand{\botF}{69.0$\pm$1.7 92.3$\pm$2.4 97.1$\pm$1.1 98.3$\pm$0.6}

\newcommand{\pcbG}{3.4$\pm$0.0 7.0$\pm$0.3 8.5$\pm$0.0 12.7$\pm$0.0}
\newcommand{\mlfnG}{3.3$\pm$0.1 6.7$\pm$0.6 8.5$\pm$0.0 13.6$\pm$0.7}
\newcommand{\haG}{3.4$\pm$0.0 7.0$\pm$0.0 8.5$\pm$0.0 12.7$\pm$0.3}
\newcommand{\resG}{3.4$\pm$0.0 7.0$\pm$0.0 8.5$\pm$0.0 12.8$\pm$0.4}
\newcommand{\mdG}{3.1$\pm$0.0 5.6$\pm$0.0 8.5$\pm$0.0 13.6$\pm$0.7}
\newcommand{\botG}{5.4$\pm$0.2 6.0$\pm$1.6 8.9$\pm$1.9 9.0$\pm$2/2}

\newcommand{\pcbH}{32.5$\pm$0.5 68.9$\pm$1.9 75.0$\pm$1.1 77.8$\pm$0.9}
\newcommand{\mlfnH}{33.5$\pm$0.3 69.7$\pm$2.5 74.9$\pm$0.6 77.5$\pm$0.8}
\newcommand{\haH}{32.0$\pm$0.4 69.6$\pm$1.8 73.9$\pm$0.7 76.1$\pm$0.0}
\newcommand{\resH}{32.4$\pm$0.6 68.7$\pm$2.4 75.4$\pm$1.0 78.0$\pm$0.9}
\newcommand{\mdH}{26.8$\pm$0.3 59.4$\pm$2.1 71.4$\pm$1.3 77.1$\pm$0.6}
\newcommand{\botH}{33.5$\pm$0.4 69.1$\pm$2.2 75.4$\pm$0.9 76.4$\pm$0.9}


\begin{table*}
\centering
  
  \begin{tabular}{llcccc}
    \toprule
    & Model & mAP & Rank1  & Rank5 & Rank10 \\
    \midrule
    1 & PCB~\cite{Sun_ECCV2018_PCB}    & 68.4$\pm$1.4 & 86.8$\pm$2.5 & 96.1$\pm$1.6 & 98.1$\pm$0.8 \\
    2 & MLFN~\cite{chang2018multi}     & \textbf{72.4$\pm$0.9} & 86.1$\pm$2.4 & 95.1$\pm$0.8 & 97.7$\pm$0.8 \\
    3 & HACNN~\cite{li2018harmonious}  & 67.6$\pm$0.6 & 89.2$\pm$2.4 & 96.0$\pm$1.0 & 97.6$\pm$0.6 \\
    4 & Resnet50Mid~\cite{yu2017devil} & 67.1$\pm$1.7 & 86.7$\pm$3.1 & 95.4$\pm$2.0 & 97.4$\pm$0.8 \\
    5 & MuDeep~\cite{qian2017multi}    & 53.8$\pm$0.6 & 81.1$\pm$2.0 & 95.8$\pm$0.9 & 97.2$\pm$0.0 \\
    6 & BoT~\cite{Luo2019BoT}          & 71.0$\pm$1.7 & \textbf{90.5$\pm$2.2} & \textbf{96.9$\pm$1.1} & \textbf{98.1$\pm$0.8} \\
  \bottomrule
\end{tabular}
    \caption{Improved re-ID performance based on the reduced galleries with time constraints. We choose galleries with timestamp $t_{gallery}$ that satisfy $t_{query} + T_{min} \leq t_{gallery} < t_{query} + T_{max}$ to construct reduced gallery sets. $T_{max}=30$ and $T_{min}=0$ here use minutes as units.}
    \label{tab:second_exp}
\end{table*}
\begin{table*}
\centering
  \begin{tabular}{llcccc}
    \toprule
    & Model & mAP & Rank1  & Rank5 & Rank10 \\
    \midrule
    1 & PCB~\cite{Sun_ECCV2018_PCB}    & 76.6$\pm$0.8 & 94.1$\pm$1.3 & 98.0$\pm$0.8 & 98.5$\pm$0.3 \\
    2 & MLFN~\cite{chang2018multi}     & \textbf{78.3$\pm$0.7} & 90.5$\pm$1.6 & 97.9$\pm$0.7 & \textbf{98.6$\pm$0.0} \\
    3 & HACNN~\cite{li2018harmonious}  & 75.2$\pm$0.7 & 93.9$\pm$0.8 & 97.7$\pm$0.7 & \textbf{98.6$\pm$0.0} \\
    4 & Resnet50Mid~\cite{yu2017devil} & 73.9$\pm$1.5 & 91.9$\pm$1.8 & 97.3$\pm$0.9 & 98.5$\pm$0.4 \\
    5 & MuDeep~\cite{qian2017multi}    & 65.8$\pm$0.6 & 91.4$\pm$1.7 & \textbf{98.5$\pm$0.3} & \textbf{98.6$\pm$0.0} \\
    6 & BoT~\cite{Luo2019BoT}          & 77.3$\pm$1.3 & \textbf{94.7$\pm$2.5} & 98.0$\pm$0.7 & 98.5$\pm$0.3 \\
  \bottomrule
\end{tabular}
    \caption{Improved re-ID performance by temporal re-ranking using a Gamma prior for $[0, 30)$.}
    \label{tab:reranking_results}
\end{table*}


\newcommand{\pcbdistA}{76.6$\pm$0.8 94.1$\pm$1.3 \textbf{98.0$\pm$0.8} \textbf{98.5$\pm$0.3}}
\newcommand{\mlfndistA}{78.3$\pm$0.7 90.5$\pm$1.6 97.9$\pm$0.7 \textbf{98.6$\pm$0.0}}
\newcommand{\hadistA}{75.2$\pm$0.7 93.9$\pm$0.8 97.7$\pm$0.7 \textbf{98.6$\pm$0.0}}
\newcommand{\resdistA}{73.9$\pm$1.5 \textbf{91.9$\pm$1.8} 97.3$\pm$0.9 \textbf{98.5$\pm$0.4}}
\newcommand{\mddistA}{61.5$\pm$0.8 80.8$\pm$1.9 95.9$\pm$1.7 \textbf{98.4$\pm$0.5}}
\newcommand{\botdistA}{77.3$\pm$1.3 94.7$\pm$2.5 \textbf{98.0$\pm$0.7} \textbf{98.5$\pm$0.3}}

\newcommand{\pcbdistB}{76.4$\pm$0.8 94.3$\pm$1.3 \textbf{98.0$\pm$0.8} \textbf{98.5$\pm$0.3}}
\newcommand{\mlfndistB}{78.3$\pm$0.7 91.7$\pm$1.8 \textbf{98.0$\pm$0.7} \textbf{98.6$\pm$0.0}}
\newcommand{\hadistB}{75.4$\pm$0.6 93.8$\pm$0.8 \textbf{97.8$\pm$0.7} \textbf{98.6$\pm$0.0}}
\newcommand{\resdistB}{74.3$\pm$1.3 91.8$\pm$1.8 97.3$\pm$0.9 \textbf{98.5$\pm$0.4}}
\newcommand{\mddistB}{62.2$\pm$0.8 81.1$\pm$2.3 96.1$\pm$1.7 \textbf{98.4$\pm$0.5}}
\newcommand{\botdistB}{77.1$\pm$1.2 95.2$\pm$2.3 97.8$\pm$0.7 \textbf{98.5$\pm$0.3}}

\newcommand{\pcbdistC}{77.1$\pm$0.8 \textbf{94.9$\pm$1.3} \textbf{98.0$\pm$0.7} \textbf{98.5$\pm$0.3}}
\newcommand{\mlfndistC}{79.1$\pm$0.8 92.0$\pm$1.6 97.8$\pm$0.7 98.5$\pm$0.3}
\newcommand{\hadistC}{75.8$\pm$0.6 93.8$\pm$0.8 \textbf{97.8$\pm$0.7} \textbf{98.6$\pm$0.0}}
\newcommand{\resdistC}{74.6$\pm$1.3 91.9$\pm$2.2 \textbf{97.5$\pm$0.8} \textbf{98.5$\pm$0.4}}
\newcommand{\mddistC}{62.1$\pm$0.8 81.1$\pm$2.5 \textbf{96.7$\pm$1.3} \textbf{98.4$\pm$0.5}}
\newcommand{\botdistC}{\textbf{77.7$\pm$1.4} 95.3$\pm$2.4 97.9$\pm$0.7 \textbf{98.5$\pm$0.3}}

\newcommand{\pcbdistD}{76.4$\pm$0.7 94.2$\pm$1.2 \textbf{98.0$\pm$0.8} \textbf{98.5$\pm$0.3}}
\newcommand{\mlfndistD}{78.2$\pm$0.7 91.1$\pm$1.6 \textbf{98.0$\pm$0.7} \textbf{98.6$\pm$0.0}}
\newcommand{\hadistD}{75.4$\pm$0.6 94.0$\pm$0.9 97.7$\pm$0.7 \textbf{98.6$\pm$0.0}}
\newcommand{\resdistD}{74.4$\pm$1.3 91.8$\pm$1.8 \textbf{97.5$\pm$0.9} \textbf{98.5$\pm$0.4}}
\newcommand{\mddistD}{62.0$\pm$0.8 80.8$\pm$2.6 96.0$\pm$1.7 \textbf{98.4$\pm$0.5}}
\newcommand{\botdistD}{77.1$\pm$1.2 95.0$\pm$2.2 \textbf{98.0$\pm$0.7} \textbf{98.5$\pm$0.3}}

\newcommand{\pcbdistE}{75.3$\pm$0.9 91.9$\pm$1.5 97.6$\pm$0.9 \textbf{98.5$\pm$0.4}}
\newcommand{\mlfndistE}{77.7$\pm$0.9 90.5$\pm$2.5 96.7$\pm$1.1 98.2$\pm$0.6}
\newcommand{\hadistE}{74.9$\pm$0.6 92.5$\pm$1.8 97.4$\pm$0.7 98.5$\pm$0.4}
\newcommand{\resdistE}{73.1$\pm$1.4 91.0$\pm$2.3 96.9$\pm$1.1 98.4$\pm$0.5}
\newcommand{\mddistE}{61.3$\pm$0.8 79.3$\pm$2.5 93.0$\pm$1.5 96.7$\pm$1.0}
\newcommand{\botdistE}{76.3$\pm$1.4 95.0$\pm$2.3 97.8$\pm$1.1 \textbf{98.5$\pm$0.4}}

\newcommand{\pcbdistF}{76.8$\pm$0.7 \textbf{94.9$\pm$1.4} \textbf{98.0$\pm$0.7} \textbf{98.5$\pm$0.3}}
\newcommand{\mlfndistF}{78.5$\pm$0.8 92.3$\pm$1.4 97.8$\pm$0.9 98.5$\pm$0.3}
\newcommand{\hadistF}{75.6$\pm$0.6 93.9$\pm$0.9 97.7$\pm$0.7 \textbf{98.6$\pm$0.0}}
\newcommand{\resdistF}{74.2$\pm$1.3 91.8$\pm$2.5 97.3$\pm$0.8 \textbf{98.5$\pm$0.4}}
\newcommand{\mddistF}{62.9$\pm$0.7 \textbf{83.2$\pm$2.5} 95.5$\pm$1.3 98.0$\pm$0.8}
\newcommand{\botdistF}{77.0$\pm$1.2 \textbf{95.6$\pm$2.5} \textbf{98.0$\pm$0.7} \textbf{98.5$\pm$0.3}}

\newcommand{\pcbdistG}{76.4$\pm$0.8 91.8$\pm$1.5 97.7$\pm$0.8 \textbf{98.5$\pm$0.3}}
\newcommand{\mlfndistG}{78.5$\pm$0.9 90.7$\pm$2.1 97.0$\pm$1.0 98.4$\pm$0.5}
\newcommand{\hadistG}{75.6$\pm$0.6 92.7$\pm$1.6 97.5$\pm$0.8 98.4$\pm$0.5}
\newcommand{\resdistG}{73.9$\pm$1.2 90.4$\pm$2.4 97.1$\pm$1.0 \textbf{98.5$\pm$0.4}}
\newcommand{\mddistG}{61.1$\pm$0.7 75.6$\pm$2.4 90.4$\pm$2.2 95.7$\pm$1.2}
\newcommand{\botdistG}{77.1$\pm$1.3 94.7$\pm$2.1 97.8$\pm$0.7 \textbf{98.5$\pm$0.4}}

\newcommand{\pcbdistH}{\textbf{77.2$\pm$0.7} \textbf{94.9$\pm$1.6} \textbf{98.0$\pm$0.7} \textbf{98.5$\pm$0.3}}
\newcommand{\mlfndistH}{\textbf{79.2$\pm$0.8} \textbf{92.2$\pm$1.5} 97.8$\pm$0.9 98.5$\pm$0.3}
\newcommand{\hadistH}{\textbf{75.9$\pm$0.6} \textbf{94.2$\pm$1.3} 97.7$\pm$0.7 \textbf{98.6$\pm$0.0}}
\newcommand{\resdistH}{\textbf{74.7$\pm$1.3} \textbf{91.9$\pm$2.2} 97.4$\pm$0.8 \textbf{98.5$\pm$0.4}}
\newcommand{\mddistH}{\textbf{62.7$\pm$0.8} 82.8$\pm$2.4 95.5$\pm$1.3 97.9$\pm$0.8}
\newcommand{\botdistH}{77.6$\pm$1.4 95.5$\pm$2.2 \textbf{98.0$\pm$0.8} \textbf{98.5$\pm$0.3}}

\newcommand{\pcbdistI}{76.7$\pm$0.8 94.4$\pm$1.1 \textbf{98.0$\pm$0.7} \textbf{98.5$\pm$0.3}}
\newcommand{\mlfndistI}{79.0$\pm$0.8 \textbf{92.2$\pm$1.5} 97.8$\pm$0.7 98.5$\pm$0.3}
\newcommand{\hadistI}{75.5$\pm$0.6 94.0$\pm$0.8 97.7$\pm$0.7 \textbf{98.6$\pm$0.0}}
\newcommand{\resdistI}{74.3$\pm$1.3 \textbf{91.9$\pm$1.8} 97.3$\pm$0.9 \textbf{98.5$\pm$0.4}}
\newcommand{\mddistI}{62.2$\pm$0.8 81.4$\pm$2.4 96.3$\pm$1.6 \textbf{98.4$\pm$0.5}}
\newcommand{\botdistI}{77.4$\pm$1.4 95.3$\pm$2.2 97.9$\pm$0.7 \textbf{98.5$\pm$0.3}}

\newcommand{\pcbdistJ}{76.3$\pm$0.9 90.9$\pm$2.0 97.7$\pm$0.8 \textbf{98.5$\pm$0.3}}
\newcommand{\mlfndistJ}{78.1$\pm$0.7 90.5$\pm$1.9 96.8$\pm$1.1 98.2$\pm$0.6}
\newcommand{\hadistJ}{75.6$\pm$0.6 92.7$\pm$1.6 97.5$\pm$0.8 98.4$\pm$0.5}
\newcommand{\resdistJ}{73.9$\pm$1.2 90.4$\pm$2.5 97.1$\pm$1.0 \textbf{98.5$\pm$0.4}}
\newcommand{\mddistJ}{61.4$\pm$0.8 77.3$\pm$1.6 92.0$\pm$2.2 96.6$\pm$1.0}
\newcommand{\botdistJ}{77.1$\pm$1.3 94.7$\pm$2.1 97.8$\pm$0.7 \textbf{98.5$\pm$0.4}}


\section{A Novel Temporal Re-Ranking Approach}
\label{sec:re-ranking}
The previous experiments of eliminating candidate images that appear outside the $\Delta t$ minute window from the query image corresponds to applying a box shaped prior distribution to the gallery images. Essentially, this sets the prior probability of an image containing the query identity to zero outside the $\Delta t$ minute window, and gives equal probability to each candidate image within the $\Delta t$ minute window. In the following, we describe experiments with more sophisticated priors which have been chosen to reflect the prior belief that, having left a camera, a person is likely to appear in another sooner rather than later. We then use the Bayesian update formula to integrate the evidence coming from the appearance models and the posterior probability of each candidate being relevant.

The following describes the theoretical motivation behind the approach. Let $Y$ be the event that a gallery image described by a feature vector $x_g$ contains a person of the same identity as the query image described by a feature vector $x_q$. Also, let $\Delta t$ be the time difference between the observation of $x_g$ and $x_q$. $\Delta s$ is the walking distance between cameras that the observation of $x_g$ and $x_q$ appear. By Bayes rule we have:
\begin{equation}
P(Y|x_q, x_g,\Delta t, \Delta s) \propto P(x_q, x_g|Y, \Delta t, \Delta s)P(Y|\Delta t, \Delta s),
\end{equation}
where $P(x_q, x_g|Y, \Delta t)$ is the probability of observing feature vectors $x_g$ and $x_q$ given that they contain the same identity and $P(Y|\Delta t)$ can be interpreted as the prior probability of relevance conditioned only on time. If we further assume that the probability of observing feature vectors $x_g$ and $x_q$ is conditionally independent of $\Delta t$ and $\Delta s$ given $Y$, and also make the naive assumption that $\Delta t$ and $\Delta s$ are also independent, then this simplifies to:
\begin{equation}
\label{prob_rerank}
P(Y|x_q, x_g,\Delta t, \Delta s) \propto P(x_q, x_g|Y)P(Y|\Delta t)P(Y|\Delta s).
\end{equation}
We model the likelihood term on the right hand side of Eq~\ref{prob_rerank} using a normal distribution centered on the query, and we use Laplace distribution to model spatial prior. That is:
\begin{equation}
P(x_q, x_g|Y)  \propto \exp \bigg{\{} - \frac{\|x_q - x_g\|^2}{2\sigma ^2} \bigg{\}},
\end{equation}
\begin{equation}
P(Y|\Delta s) \propto \exp \bigg{\{} - \frac{\|\Delta s\|}{\sigma_s} \bigg{\}},
\end{equation}
where $\sigma$ is the standard deviation of the Gaussian and controls how quickly the probability decays as the gallery image moves away from the query image in feature space. Larger $\sigma$ values reduce the effect of the time prior. $\sigma$ needs to be re-calibrated for each model separately to account for the differing scales of the feature space. The detailed choice of $\sigma$ for Table~\ref{tab:reranking_results} is reported in Table~\ref{tab:sigma}.

\begin{figure}[h]
  \centering
  \includegraphics[width=\linewidth]{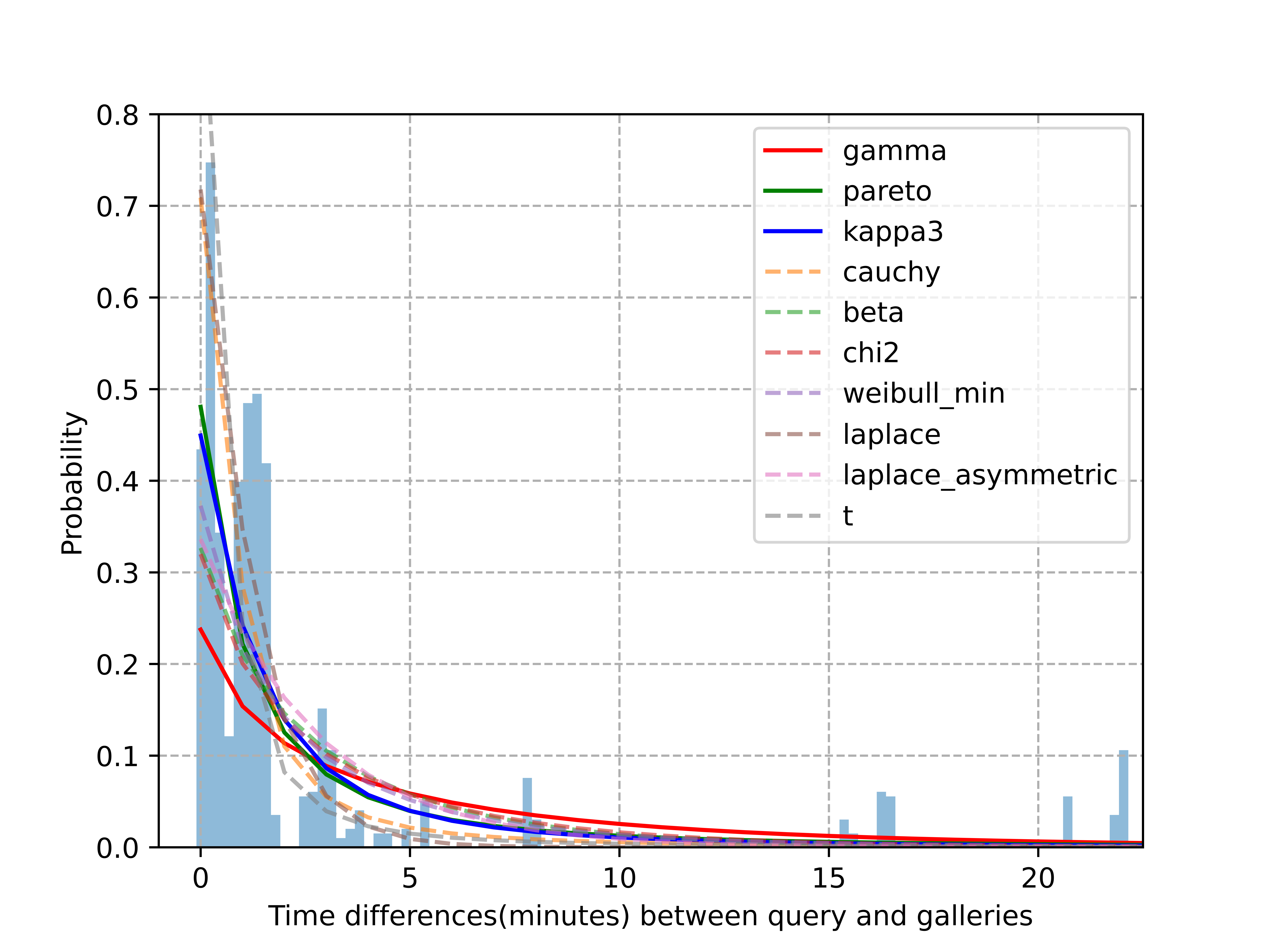}
  \caption{Empirical Distribution of $P(Y|\Delta t)$ computed from the DAA dataset. We fit various probability density functions to the empirical distribution and use them for time constrained re-ranking.}
  \label{fig:gamma}
\end{figure}

The prior term $P(Y|\Delta t)$ in Eq~\ref{prob_rerank} should be selected to reflect the fact that having left a camera, the query person is more likely to appear in another camera sooner rather than later. For example we could choose the Gamma family of distributions to allow flexible control over the prior. In this case, we model the temporal prior term using:
\begin{equation}
P(Y|\Delta t) \propto (\Delta t)^{\alpha - 1}e^{-\Delta t/\beta},
\end{equation}
where $\alpha$ and $\beta$ are hyperparameters controlling the shape of the distribution. Fig.~\ref{fig:gamma} shows various probability distributions of which the parameters are estimated using empirical distribution of the DAA dataset. Table~\ref{tab:distribution_param} gives the parameters fit for each probability distribution. Notice how the distributions place more probability that relevant images will appear earlier, and that they can be configured to account for a delay between leaving one camera and entering another.
\begin{figure*}
  \centering
  \includegraphics[width=5.5in]{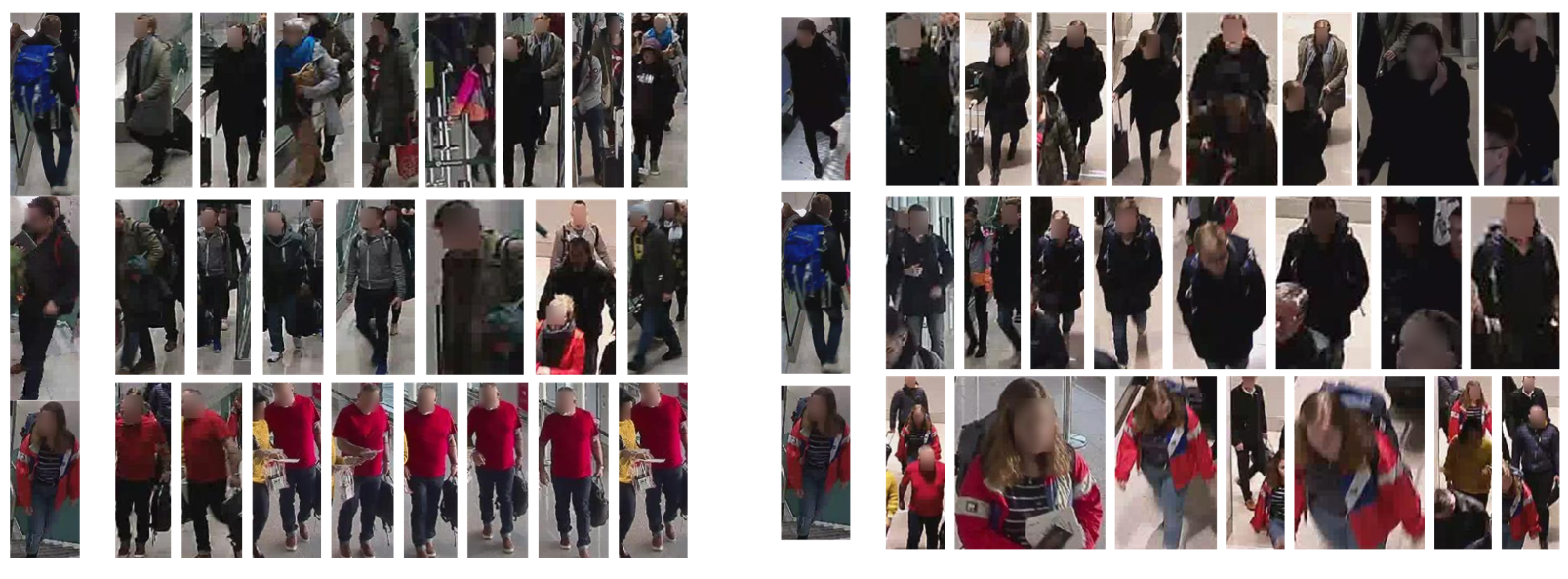}
  \caption{Examples of query images and retrieved false positive (left) and false negative (right) samples. The left columns are the query images and the remaining images are retrieved examples.}
  \label{fig:fn_example}
\end{figure*}
\begin{table*}
\centering
{%
  \begin{tabular}{llcccccccc}
    \toprule
    &  &\multicolumn{2}{c}{$\sigma_s=50$}&\multicolumn{2}{c}{$\sigma_s=100$}&\multicolumn{2}{c}{$\sigma_s=400$} & \multicolumn{2}{c}{$\Delta s$*} \\
    & Model & mAP & Rank1 & mAP & Rank1 & mAP & Rank1  & mAP & Rank1 \\
    \midrule
    1 & PCB~\cite{Sun_ECCV2018_PCB}    &$-$1.2 &$+$1.1 &$-$0.6 & $+$0.7 &-0.2&+0.3&$+$0.4 &$-$0.9 \\
    2 & MLFN~\cite{chang2018multi}     &$-$1.2 &$+$1.3 &$-$0.5 & $+$1.0 &-0.2&-0.1&$+$0.8 &$-$0.4 \\
    3 & HACNN~\cite{li2018harmonious}  &$-$1.0 &$-$0.4 &$-$0.5 & $+$0.0 &-0.1&-0.1&$+$0.4 &$-$0.2 \\
    4 & Resnet50Mid~\cite{yu2017devil} &$-$0.9 &$-$0.4 &$-$0.4 & $+$0.0 &-0.1&-0.1&$+$0.2 &$+$0.1 \\
    5 & MuDeep~\cite{qian2017multi}    &$-$2.2 &$-$0.8 &$-$1.1 & $-$0.1 &-0.3&-0.1&$+$0.8 &$+$0.0 \\
  \bottomrule
\end{tabular}}
    \caption{Improved re-ID performance by temporal and spatial re-ranking using a Gamma prior for $[0, 30)$. $\Delta s$* is using $P(Y|\Delta s)  \propto \Delta s$ as spatial prior.}
    \label{tab:reranking_spatial_temporal}
\end{table*}

After ranking by the updated posterior scores we obtain a significant improvement of \textbf{10.03}\% in mAP and \textbf{9.95}\% in Rank1 accuracy metrics compared to the results in Table~\ref{tab:second_exp}. Again, we report the average performance gain across all models. Table~\ref{tab:reranking_results} shows the performance improvements for the Gamma time prior distribution for the time interval $[0, 30)$.

An experiment using both spatial and temporal priors is reported in Table~\ref{tab:reranking_spatial_temporal}. We can see that comparing with temporal re-ranking alone, the result is almost identical. There are two possible explanations. First, the spatial information between subjects in the gallery and probe images are estimated using walking distances between cameras (see Table~\ref{tab:walkingdistances}) and this estimation largely ignores subtle spatial differences between subjects. 
\begin{table}[ht]
\centering
  \begin{tabular}{llllll}
    \toprule
     & \textbf{C900} & \textbf{C902} & \textbf{C903}  & \textbf{C904} & \textbf{C926} \\
    \midrule
    \textbf{C900} & 0.0 & 48.5 & 106.0 & 70.0 & 68.0 \\
    \textbf{C902} & 48.5 & 0.0 & 59.0 & 19.0 & 38.5 \\
    \textbf{C903} & 106.0 & 59.0 & 0.0 & 45.0 & 97.5 \\
    \textbf{C904} & 70.0 & 19.0 & 45.0 & 0.0 & 63.5 \\
    \textbf{C926} & 68.0 & 38.5 & 97.5 & 57.0 & 0.0 \\
    \bottomrule
  \end{tabular}
  \caption{Estimated relative walking distances between cameras.}
  \label{tab:walkingdistances}
\end{table}
Second, the temporal and spatial priors are related since the more time passed, the larger the chance a subject could have walked further. Introducing the spatial prior provides little new information to improve the ranking.
Our performance gain is on a par with the results reported in \cite{Zhong_CVPR2017}, which are based on the commonly used but computationally expensive re-ranking approach using $k$-reciprocal encoding. 

\begin{table*}
\centering
  \setlength\tabcolsep{3pt}
  \small
  \begin{tabular}{lcccccccccc}
    \toprule
     & Gamma & Beta &Pareto& $\chi^2$ & Laplace & \makecell{Laplace-\\asymm}&$t$& Kappa(3) & \makecell{Weibull-\\min} & Cauchy \\ \midrule
    param &\makecell{$a$\\0.5} &\makecell{$a, b$ \\$0.68,98.68$} & \makecell{$b$\\1.2} & \makecell{df\\1.19} & N/A &\makecell{$\kappa$\\0.1} &\makecell{df\\1} &\makecell{$a$\\1.5} &\makecell{$c$\\0.74}&N/A\\ \midrule
    loc        & -0.1  & -0.1        & -1.4 & -0.1 & 1.1 & 0 & 0.9   &-0.1 &-0.1 & 0.9\\
    scale      & 9.4   & 416.75      & 1.3 & 2.46 & 2.3 & 0.3 & 0.63 &1.6 &2.2 & 0.633\\
    \bottomrule
  \end{tabular}
  \caption{Estimated parameters of empirical distributions.}
  \label{tab:distribution_param}
\end{table*}
              
\begin{table*}
\centering
  \setlength\tabcolsep{3pt}
  \small
  \begin{tabular}{lcccccccccc}
    \toprule
               & Gamma & Beta        &Pareto& $\chi^2$ & Laplace & \makecell{Laplace-\\asymm}&$t$       & Kappa(3) & \makecell{Weibull-\\min} & Cauchy \\ \midrule
    PCB        &1.6    &1.8          &1.4   &1.7   &1.2      &2                              &1.1     &1.6     &1.6          &1.3                \\
    MLFN       &80     &80           &60    &80    &60       &80                             &50      &60      &60           &60                 \\
    HACNN      &0.6    &0.7          &0.6   &0.6   &0.7      &0.8                            &0.5     &0.6     &0.6          &0.5                 \\ 
    Resnet50mid&150    &160          &150   &170   &150      &180                            &150     &160     &150          &150                 \\ 
    MuDeep     &450   &500         &500  &500  &450     &700                           &400    &500    &500         &500                 \\
    BoT        & 0.4    & 0.4    &   0.3   &   0.4   &     0.3    &      0.5         & 0.3       & 0.3    & 0.3           & 0.3                \\
    \bottomrule
  \end{tabular}
  \caption{Selected $\sigma$ for different models.}
  \label{tab:sigma}
\end{table*}
\section{Discussion}
\label{sec:discussion}


The individuals annotated and re-identified in this study were typically wearing brightly colored or distinctive clothes, meaning that model performance would likely degrade with more varied query individuals, as even humans find it harder to distinguish between people wearing darker clothing in CCTV images.

The proposed approach, which takes full advantage of time information, could greatly improve the number of true positive samples retrieved. In fact, by applying time constraints, it increases the number of true positives by around 200\%. Applying both prior and time constraints, it increases by around 210\%, while both false positives and negatives are suppressed. There are still some hard samples in the gallery that are difficult to separate in the feature space.  Figure~\ref{fig:fn_example} illustrates some false negative and false positive examples after applying the prior and time constraints. To determine the cut-off rank that dictates the number of negative and positive examples, we use a different ranking threshold for each query. For instance, if there should be 10 matched IDs in the gallery for a certain query, the cut-off rank would be 10, meaning that the top-10 images are retrieved as positive and the remainder is marked negative. We noticed that even with the introduction of time information, pedestrian pose changes and similar clothing colors and styles could be contributing factors to the increasing false positives and negatives.

Notwithstanding, the ability to successfully re-identify at least a proportion of people from annotated images derived from CCTV footage, as demonstrated in this paper, could be of significant interest to the airport authorities. If we assume that the majority of people will move through the airport in a similar way, then the combination of two approaches: re-ID for a few individuals within a crowd and crowd density mapping, could provide key insights into passenger flow. For example, being able to evaluate whether an area of high crowd density at a given point in time is likely to dissipate or transfer to another area in the airport.

The proposed dataset with time prior re-ranking could also be of benefit to the recent development of unsupervised and/or self-supervised learning \cite{UnReID1_CVPRW2019,UnReID2_ICCV2019} in re-ID research since the re-ranking process is simple, efficient, and fast. Given timestamp information, re-ranking could plug into any of the current implementations of unsupervised and/or self-supervised learning pipelines.

\section{Conclusion}

Most of the published person re-ID algorithms conduct supervised training and
testing on single labeled datasets of small size, so directly deploying these trained models to a large-scale real-world camera network may lead to poor performance due to underfitting \cite{Lv_CVPR2018}. Also, most re-ID algorithms consider only appearance models and thus concentrate on deep learning for visual feature representation \cite{Wang_AAAI2019}. Our initial experiments with additional temporal constraints demonstrate considerable improvement in person re-ID performance on our DAA dataset, which is now publicly available. This research is not possible on other image-based datasets. Using time-reduced galleries and additional temporal re-ranking step yields the significant performance gain of \textbf{47.46}\% in mAP and \textbf{40.17}\% in Rank1 accuracy as compared to the results previously reported.

\section*{Acknowledgments}

This publication has emanated from research supported by Science Foundation Ireland (SFI) under grant numbers SFI/20/COV/8579 SFI/12/RC/2289\_2, co-funded by the European Regional Development Fund. We would like to thank Dublin Airport Authority for their contributions to this work.

{\small
\bibliographystyle{ieee_fullname}
\bibliography{WACV22_main_paper}

\begin{thebibliography}{10}\itemsep=-1pt

\bibitem{Camps_IEEECSVT2017}
Octavia Camps, Mengran Gou, Tom Hebble, Srikrishna Karanam, Oliver Lehmann,
  Yang Li, Richard~J. Radke, Ziyan Wu, and Fei Xiong.
\newblock From the lab to the real world: Re-identification in an airport
  camera network.
\newblock {\em IEEE Transactions on Circuits and Systems for Video Technology},
  27(3), 2017.

\bibitem{chang2018multi}
Xiaobin Chang, Timothy~M Hospedales, and Tao Xiang.
\newblock Multi-level factorisation net for person re-identification.
\newblock In {\em Proceedings of the IEEE Conference on Computer Vision and
  Pattern Recognition}, pages 2109--2118, 2018.

\bibitem{Cho_CVIU2019}
Yeong-Jun Cho, Su-A Kim, Jae-Han Park, Kyuewang Lee, and Kuk-Jin Yoon.
\newblock Joint person re-identification and camera network topology inference
  in multiple cameras.
\newblock {\em Computer Vision and Image Understanding}, 180:34--46, 2019.

\bibitem{Dietlmeier_ICPR2020}
Julia Dietlmeier, Joseph Antony, Kevin McGuinness, and Noel~E. O’Connor.
\newblock How important are faces for person re-identification?
\newblock In {\em Proceedings of the 25th International Conference on Pattern
  Recognition}, 2021.

\bibitem{Dollar_TPAMI2014_ACF}
Piotr Dollár, Ron Appel, Serge Belongie, and Pietro Perona.
\newblock Fast feature pyramids for object detection.
\newblock {\em IEEE Transactions on Pattern Analysis and Machine Intelligence},
  36(8):1532–1545, 2014.

\bibitem{Felzenszwalb_TPAMI2010_DPM}
Pedro~F. Felzenszwalb, Ross~B. Girshick, David McAllester, and Deva Ramanan.
\newblock Object detection with discriminatively trained part-based models.
\newblock {\em IEEE Transactions on Pattern Analysis and Machine Intelligence},
  32(9):1627–1645, 2010.

\bibitem{Gou_CVPR2017_DukeMTMC}
Mengran Gou, Srikrishna Karanam, Wenqian Liu, Octavia Camps, and Richard~J.
  Radke.
\newblock Dukemtmc4reid: A large-scale multi-camera person re-identification
  dataset.
\newblock In {\em CVPR Workshop}, pages 10--19, 2017.

\bibitem{Gray_ECCV2008_ViPER}
Douglas Gray and Hai Tao.
\newblock Viewpoint invariant pedestrian recognition with an ensemble of
  localized features.
\newblock In {\em Proceedings of the European Conference on Computer Vision},
  2008.

\bibitem{Hu_CVPR2017_TinyFaces}
Peiyun Hu and Deva Ramanan.
\newblock Finding tiny faces.
\newblock In {\em Proceedings of the IEEE Conference on Computer Vision and
  Pattern Recognition}, pages 951--959, 2017.

\bibitem{Karanam_IEEETPAMI2018}
Srikrishna Karanam, Mengran Gou, Ziyan Wu, Angels Rates-Borras, Octavia Camps,
  and Richard~J. Radke.
\newblock A systematic evaluation and benchmark for person re-identification:
  Features, metrics, and datasets.
\newblock {\em IEEE Transactions on Pattern Analysis and Machine Intelligence},
  41(3):523--536, 2018.

\bibitem{Li_CVPR2014_CUHK03}
Wei Li, Rui Zhao, Tong Xiao, and Xiaogang Wang.
\newblock Deepreid: Deep filter pairing neural network for person
  re-identification.
\newblock In {\em Proceedings of the IEEE Conference on Computer Vision and
  Pattern Recognition}, 2014.

\bibitem{li2018harmonious}
Wei Li, Xiatian Zhu, and Shaogang Gong.
\newblock Harmonious attention network for person re-identification.
\newblock In {\em Proceedings of the IEEE conference on computer vision and
  pattern recognition}, pages 2285--2294, 2018.

\bibitem{UnReID2_ICCV2019}
Yu-Jhe Li, Ci-Siang Lin, Yan-Bo Lin, and Yu-Chiang~Frank Wang.
\newblock Cross-dataset person re-identification via unsupervised pose
  disentanglement and adaptation.
\newblock In {\em ICCV}, 2019.

\bibitem{Li_2020_STreranking}
Zhongyi Li, Yi Jin, Yidong Li, Congyan Lang, Songhe Feng, and Tao Wang.
\newblock Learning part-alignment feature for person re-identification with
  spatial-temporal-based re-ranking method.
\newblock {\em World Wide Web (Springer)}, 23:1907--1923, 2020.

\bibitem{liao2020interpretable}
Shengcai Liao and Ling Shao.
\newblock Interpretable and generalizable person re-identification with
  query-adaptive convolution and temporal lifting.
\newblock In {\em Computer Vision--ECCV 2020: 16th European Conference,
  Glasgow, UK, August 23--28, 2020, Proceedings, Part XI 16}, pages 456--474.
  Springer, 2020.

\bibitem{Liu_ECCV2016_SSD}
Wei Liu, Dragomir Anguelov, Dumitru Erhan, Christian Szegedy, Scott Reed,
  Cheng-Yang Fu, and Alexander~C. Berg.
\newblock Ssd: Single shot multibox detector.
\newblock In {\em Proceedings of the European Conference on Computer Vision},
  2016.

\bibitem{GRID_CVPR2009}
Chen~Change Loy, Tao Xiang, and Shaogang Gong.
\newblock Multi-camera activity correlation analysis.
\newblock In {\em Proceedings of the IEEE Conference on Computer Vision and
  Pattern Recognition}, 2009.

\bibitem{Luo2019BoT}
H. {Luo}, W. {Jiang}, Y. {Gu}, F. {Liu}, X. {Liao}, S. {Lai}, and J. {Gu}.
\newblock A strong baseline and batch normalization neck for deep person
  re-identification.
\newblock {\em IEEE Transactions on Multimedia}, 22(10):2597--2609, 2020.

\bibitem{Lv_CVPR2018}
Jianming Lv, Weihang Chen, Qing Li, and Can Yang.
\newblock Unsupervised cross-dataset person re-identification by transfer
  learning of spatial-temporal patterns.
\newblock In {\em Proceedings of the IEEE Conference on Computer Vision and
  Pattern Recognition}, pages 7948--7956, 2018.

\bibitem{malin2011never}
Bradley Malin, Kathleen Benitez, and Daniel Masys.
\newblock Never too old for anonymity: a statistical standard for demographic
  data sharing via the hipaa privacy rule.
\newblock {\em Journal of the American Medical Informatics Association},
  18(1):3--10, 2011.

\bibitem{Meng_CVPR2019_WeaklyReID}
Jingke Meng, Sheng Wu, and Wei-Shi Zheng.
\newblock Weakly supervised person re-identification.
\newblock In {\em Proceedings of the IEEE Conference on Computer Vision and
  Pattern Recognition}, pages 760--769, 2019.

\bibitem{Murgia_FT2019}
Madhumita Murgia.
\newblock Who's using your face? the ugly truth about facial recognition.
\newblock {\em Financial Times}, 2019.

\bibitem{qian2017multi}
Xuelin Qian, Yanwei Fu, Yu-Gang Jiang, Tao Xiang, and Xiangyang Xue.
\newblock Multi-scale deep learning architectures for person re-identification.
\newblock In {\em Proceedings of the IEEE International Conference on Computer
  Vision}, pages 5399--5408, 2017.

\bibitem{Sun_ECCV2018_PCB}
Yifan Sun, Liang Zheng, Yi Yang, Qi Tian, and Shengjin Wang.
\newblock Beyond part models: Person retrieval with refined part pooling (and a
  strong convolutional baseline.
\newblock In {\em Proceedings of the European Conference on Computer Vision},
  2018.

\bibitem{UnReID1_CVPRW2019}
Haotian Tang, Yiru Zhao, and Hongtao Lu.
\newblock Unsupervised person re-identification with iterative self-supervised
  domain adaptation.
\newblock In {\em CVPRW}, 2019.

\bibitem{Wang_AAAI2019}
Guangcong Wang, Jianhuang Lai, Peigen Huang, and Xiaohua Xie.
\newblock Spatial-temporal person re-identification.
\newblock In {\em Proceedings of the AAAI Conference on Artificial
  Intelligence}, volume~33, pages 8933--8940, 2019.

\bibitem{Ye_SurveyTPAMI2021}
Mang Ye, Jianbing Shen, Gaojie Lin, Tao Xiang, Ling Shao, and Steven~C.~H. Hoi.
\newblock Deep learning for person re-identification: A survey and outlook.
\newblock {\em IEEE Transactions on Pattern Analysis and Machine Intelligence},
  2021.

\bibitem{yu2017devil}
Qian Yu, Xiaobin Chang, Yi-Zhe Song, Tao Xiang, and Timothy~M Hospedales.
\newblock The devil is in the middle: Exploiting mid-level representations for
  cross-domain instance matching.
\newblock {\em arXiv preprint arXiv:1711.08106}, 2017.

\bibitem{Zheng_ICCV2015_Market1501}
Liang Zheng, Liyue Shen, Lu Tian, Shengjin Wang, Jingdong Wang, and Qi Tian.
\newblock Scalable person re-identification: A benchmark.
\newblock In {\em Proceedings of the IEEE International Conference on Computer
  Vision}, 2015.

\bibitem{Zheng_PastPresentFuture_arXiv2016}
Liang Zheng, Yi Yang, and Alexander~G. Hauptmann.
\newblock Person re-identification: Past, present and future.
\newblock {\em arXiv}, 2016.

\bibitem{Zheng_DukeMTMC_arXiv2017}
Zhedong Zheng, Liang Zheng, and Yi Yang.
\newblock Unlabeled samples generated by gan improve the person
  re-identification baseline in vitro.
\newblock {\em arXiv}, 2017.

\bibitem{Zhong_CVPR2017}
Zhun Zhong, Liang Zheng, Donglin Cao, and Shaozi Li.
\newblock Re-ranking person re-identification with k-reciprocal encoding.
\newblock In {\em Proceedings of the IEEE Conference on Computer Vision and
  Pattern Recognition}, pages 1318--1327, 2017.

\bibitem{KaiyangZhou_TorchReID_arXiv2019}
Kaiyang Zhou and Tao Xiang.
\newblock Torchreid: A library for deep learning person re-identification in
  pytorch.
\newblock {\em arXiv}, 2019.

\end{thebibliography}
}

\end{document}


\maketitle

\section{Camera Positions}
\begin{figure}[ht]
    \subfigure[Non-overlapping views from cameras used in our dataset.\label{fig:cameras}] {\includegraphics[width=0.45\linewidth]{cameras_blinded.png}}
    \hfill
    \subfigure[Camera topology in our XXX dataset\label{fig:airport_map}]{\includegraphics[width=0.5\linewidth]{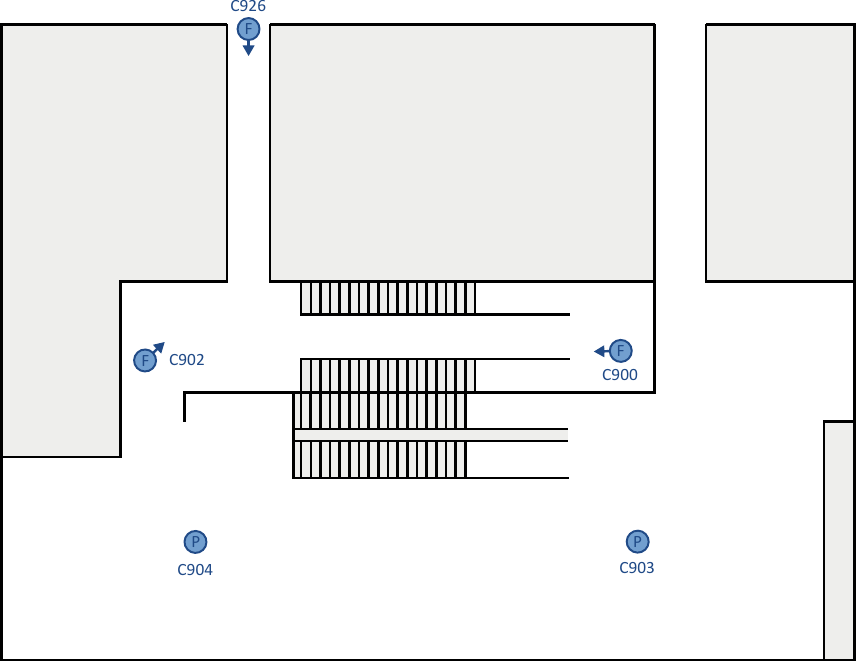}}
\caption{Camera topology and views used to create dataset}
\label{fig:topo_and_map}
\end{figure}

Figure \ref{fig:cameras} shows snapshots from the camera views used during the collection of the new dataset. Figure~\ref{fig:airport_map} provides the camera topology. The estimated relative distances and relative walking distances between cameras are provided in Tables \ref{tab:distances} and \ref{tab:walkingdistances}.

\begin{table}[ht]
\centering
  
  \begin{tabular}{llllll}
    \toprule
     & \textbf{C900} & \textbf{C902} & \textbf{C903}  & \textbf{C904} & \textbf{C926} \\
    \midrule
    \textbf{C900} & 0.0 & 48.5 & 20.0 & 47.0 & 49.5 \\
    \textbf{C902} & 48.5 & 0.0 & 52.5 & 19.0 & 19.0 \\
    \textbf{C903} & 20.0 & 52.5 & 0.0 & 45.0 & 65.5 \\
    \textbf{C904} & 47.0 & 19.0 & 45.0 & 0.0 & 51.5 \\
    \textbf{C926} & 49.5 & 35.0 & 65.5 & 51.5 & 0.0 \\
    \bottomrule
  \end{tabular}
  \caption{Estimated relative distances between cameras "as the crow files".}
  \label{tab:distances}
\end{table}

\begin{table}[ht]
\centering
  
  \begin{tabular}{llllll}
    \toprule
     & \textbf{C900} & \textbf{C902} & \textbf{C903}  & \textbf{C904} & \textbf{C926} \\
    \midrule
    \textbf{C900} & 0.0 & 48.5 & 106.0 & 70.0 & 68.0 \\
    \textbf{C902} & 48.5 & 0.0 & 59.0 & 19.0 & 38.5 \\
    \textbf{C903} & 106.0 & 59.0 & 0.0 & 45.0 & 97.5 \\
    \textbf{C904} & 70.0 & 19.0 & 45.0 & 0.0 & 63.5 \\
    \textbf{C926} & 68.0 & 38.5 & 97.5 & 57.0 & 0.0 \\
    \bottomrule
  \end{tabular}
  \caption{Estimated relative walking distances between cameras.}
  \label{tab:walkingdistances}
\end{table}

\section{Training Methodology}

In this paper we adopt the cross-dataset training methodology. In order to find out which dataset to use for training, we perform the following experiments: we train the PCB model for 60 epochs separately on the Airport and Market1501 datasets. Table~\ref{tab:first_exp} contains the results of these experiments. In particular, the PCB model was trained on 3,493 images from the training part of the Airport dataset and 12,937 images from the training part of the large-scale Market1501. The best performance was achieved by training the PCB model on the Airport dataset \cite{Camps_IEEECSVT2017,Karanam_IEEETPAMI2018}, which is to be expected given the large size of the dataset and similarity of image type when compared to our XXX dataset. In this case, the top ranked gallery image contained the query individual 67.6\% of the time, increasing to 81.7\% when the top-5 ranked gallery images were considered. After an initial ranking list is obtained, good practice consists of adding a re-ranking step, with the expectation that the relevant images will receive higher ranks. However, in this case, the additional state-of-the-art re-ranking ($k$-reciprocal encoding) procedure only improves the mAP and Rank20 metrics.

\begin{table}[ht]
\centering
  
  \begin{tabular}{lccc}
    \toprule
    Source dataset/ & Airport & Airport with  & Market1501\\
    Metrics                &         & re-ranking \cite{Zhong_CVPR2017} & \\
    \midrule
    mAP    & 32.3\% & \bf{36.3}\% & 26.8\% \\
    Rank1  & \bf{67.6}\% & 59.2\% & 49.3\% \\
    Rank5  & \bf{81.7}\% & 80.3\% & 73.2\% \\
    Rank10 & \bf{87.3}\% & 84.5\% & 80.3\% \\
    Rank20 & 88.7\% & \bf{91.5}\% & 87.3\% \\
  \bottomrule
\end{tabular}
\caption{XXX re-ID results when the PCB model was trained on different source public datasets. The best result for each metric is highlighted in bold text.}
  \label{tab:first_exp}
\end{table}

\section{Additional experimental results}
Additional experimental results are shown in Figure~\ref{fig:exp1_table} and Figure~\ref{fig:exp2_table}.
\begin{figure}[!htbp]
  \centering
  \includegraphics[width=\linewidth]{exp1_table_update.png}
  \caption{Improved re-ID performance based on the reduced galleries with time constrains. The same models as in main document are used to report means and standard deviations. $[T_{min}, T_{max})$ defines the interval from which we choose galleries. We choose galleries with timestamp $t_{gallery}$ that satisfy $t_{query} + T_{min} \leq t_{gallery} < t_{query} + T_{max}$ to construct reduced gallery sets. $T_{max}$ and $T_{min}$ here use minutes as units. For each model four metrics are reported they are mAP, Rank1, Rank5, and Rank10 corresponding to rows from top to bottom. Best results for each model are shown in bold font.}
  \label{fig:exp1_table}
\end{figure}

\begin{figure}[ht]
  \centering
  \includegraphics[width=\linewidth]{exp2_table_update.png}
  \caption{Improved re-ID performance by re-ranking using different time prior distributions for the $[0, 30)$ time interval. The same models as reported in main document are used to report means and standard deviations. For each model four metrics are reported they are mAP, Rank1, Rank5, and Rank10 corresponding to rows from top to bottom. Best results for each model are shown in bold font.}
  \label{fig:exp2_table}
\end{figure}

\section{Time prior distribution parameters}
We fit several prior distributions of empirical distribution using Python \textit{scikitlearn-stats} package. The parameters are presented in Table~\ref{tab:distribution_param}. Since features from different models have different scales, we employ different $\sigma$ for each model and distribution settings. The choice of $\sigma$ is demonstrated in Table~\ref{tab:sigma}.

\begin{table}[ht]
\centering
  
  \setlength\tabcolsep{3pt}
  \small
  \begin{tabular}{lcccccccccc}
    \toprule
     & gamma & beta &pareto& chi2 & laplace & \makecell{laplace-\\asymm}&t& kappa3 & \makecell{weibull-\\min} & cauchy \\ \midrule
    param &\makecell{$a$\\0.5} &\makecell{$a, b$ \\$0.68,98.68$} & \makecell{$b$\\1.2} & \makecell{$df$\\1.19} & N/A &\makecell{$kappa$\\0.1} &\makecell{$df$\\1} &\makecell{$a$\\1.5} &\makecell{$c$\\0.74}&N/A\\ \midrule
    loc        & -0.1  & -0.1        & -1.4 & -0.1 & 1.1 & 0 & 0.9   &-0.1 &-0.1 & 0.9\\
    scale      & 9.4   & 416.75      & 1.3 & 2.46 & 2.3 & 0.3 & 0.63 &1.6 &2.2 & 0.633\\
    \bottomrule
  \end{tabular}
  \caption{Estimated parameters of empirical distributions.}
  \label{tab:distribution_param}
\end{table}
              
\begin{table}[ht]
\centering
  
  \setlength\tabcolsep{3pt}
  \small
  \begin{tabular}{lcccccccccc}
    \toprule
               & gamma & beta        &pareto& chi2 & laplace & \makecell{laplace-\\asymm}&t       & kappa3 & \makecell{weibull-\\min} & cauchy \\ \midrule
    PCB        &1.6    &1.8          &1.4   &1.7   &1.2      &2                              &1.1     &1.6     &1.6          &1.3                \\
    MLFN       &80     &80           &60    &80    &60       &80                             &50      &60      &60           &60                 \\
    HACNN      &0.6    &0.7          &0.6   &0.6   &0.7      &0.8                            &0.5     &0.6     &0.6          &0.5                 \\ 
    Resnet50mid&150    &160          &150   &170   &150      &180                            &150     &160     &150          &150                 \\ 
    MuDeep     &450   &500         &500  &500  &450     &700                           &400    &500    &500         &500                 \\
    BoT        & 0.4    & 0.4    &   0.3   &   0.4   &     0.3    &      0.5         & 0.3       & 0.3    & 0.3           & 0.3                \\
    \bottomrule
  \end{tabular}
  \caption{Selected $\sigma$ for different models.}
  \label{tab:sigma}
\end{table}

\section{Additional Visualizations}
\begin{figure}[ht]
  \centering
  \includegraphics[width=0.7\linewidth]{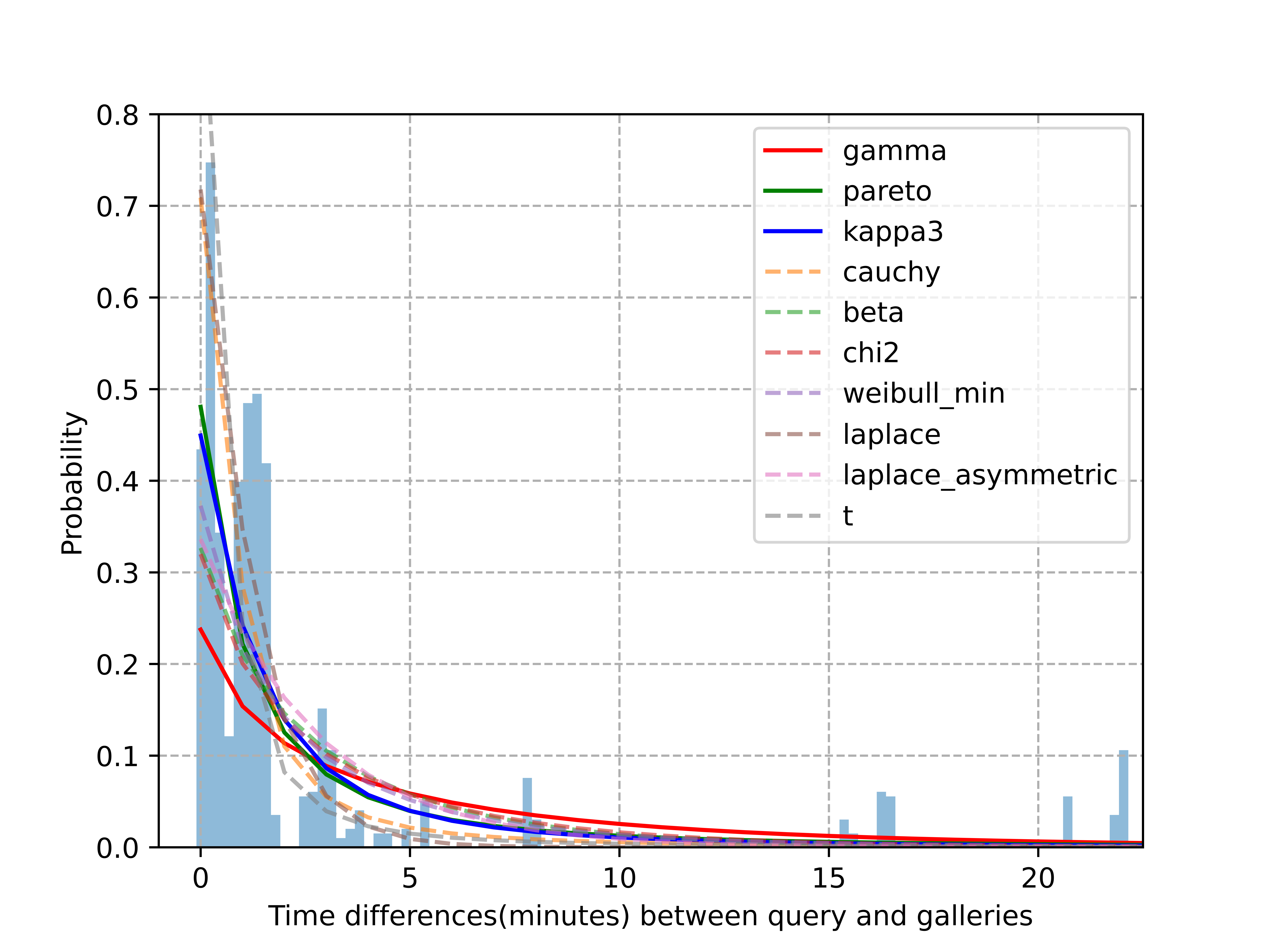}
  \caption{Empirical Distribution of $P(Y|\Delta t)$ computed from XXX dataset. We fit various probability density functions to the empirical distribution and use them for time constrained re-ranking as described in the main body of the paper.}
  \label{fig:gamma}
\end{figure}

The UMAP visualization in Figure \ref{fig:umap} shows the gallery image features extracted by the PCB model. It demonstrates diversity of distinctive clothing among identities in the dataset. Figure \ref{fig:gamma} shows the underlying distribution of time differences between query and gallery identities as well as how the various distributions fit to this data.

\begin{figure}[ht]
  \centering
  \includegraphics[width=0.6\linewidth]{umap_daa.png}
  \caption{The UMAP visualization \cite{UMAP_arXiv2020} of the $1029 \times 12288$ PCB-based (trained for 60 epochs on the Airport dataset) gallery features of the XXX dataset. It shows the diversity of distinctive clothes, which are clustered together by color.}
  \label{fig:umap}
\end{figure}

\nobibliography{ACMMM2021}